\documentclass{article}
\usepackage{amsmath,epsfig}
\usepackage[preprint]{camera_ready/spconfa4}

\usepackage{hyperref}
\usepackage{enumitem}
\usepackage{subfigure}
\usepackage{amsfonts}
\usepackage{amssymb}
\usepackage{amsthm,amsmath}
\usepackage{mathrsfs}
\usepackage{indentfirst}
\usepackage{multirow}
\usepackage{newfloat}
\usepackage{extarrows}
\usepackage{newfloat}
\usepackage{caption}
\usepackage{booktabs}
\let\OLDthebibliography\thebibliography
\renewcommand\thebibliography[1]{
  \OLDthebibliography{#1}
  \setlength{\parskip}{0pt}
  \setlength{\itemsep}{0pt plus 0.3ex}
}

\begin{document}\sloppy

\def\x{{\mathbf x}}
\def\L{{\cal L}}

\title{CMS-LSTM: Context Embedding and Multi-Scale Spatiotemporal Expression LSTM for Predictive Learning}

\name{Zenghao Chai$^1$,
Zhengzhuo Xu$^1$,
Yunpeng Bai$^1$,
Zhihui Lin$^1$,
Chun Yuan$^{1,2}$}
\address{
$^{1}$Shenzhen International Graduate School, Tsinghua University, $^{2}$Peng Cheng Laboratory\\
zenghaochai@gmail.com, 
\{xzz20, byp20, lin-zh14\}@tsinghua.edu.cn,
yuanc@sz.tsinghua.edu.cn
}

\maketitle

\begin{abstract}
Spatiotemporal predictive learning (ST-PL) is a hotspot with numerous applications, such as object movement and meteorological prediction. It aims at predicting the subsequent frames via observed sequences. However, inherent uncertainty among consecutive frames exacerbates the difficulty in long-term prediction. To tackle the increasing ambiguity during forecasting, we design CMS-LSTM to focus on context correlations and multi-scale spatiotemporal flow with details on fine-grained locals, containing two elaborate designed blocks: Context Embedding (CE) and Spatiotemporal Expression (SE) blocks. CE is designed for abundant context interactions, while SE focuses on multi-scale spatiotemporal expression in hidden states. The newly introduced blocks also facilitate other spatiotemporal models (e.g., PredRNN, SA-ConvLSTM) to produce representative implicit features for ST-PL and improve prediction quality. Qualitative and quantitative experiments demonstrate the effectiveness and flexibility of our proposed method. With fewer params, CMS-LSTM outperforms state-of-the-art methods in numbers of metrics on two representative benchmarks and scenarios. Code is available at \href{https://github.com/czh-98/CMS-LSTM}{https://github.com/czh-98/CMS-LSTM}.
\end{abstract}

\section{Introduction}

\textbf{S}patio-\textbf{T}emporal \textbf{P}redictive \textbf{L}earning (ST-PL) is one of the hotspots in predictive learning with broad research prospects in computer vision. 
The core task and challenge are predicting future sequences based on limited observed frames, containing a large amount of visual information and profound dynamic changes. Recent years have seen significant progress in ST-PL. Numerous researchers have carried out in-depth research and proposed a series of RNNs~\cite{hochreiter1997long} (especially LSTMs) based models, from the original ConvLSTM~\cite{shi2015convolutional,DBLP:conf/nips/ShiGL0YWW17} used for precipitation nowcasting to the improved approaches, such as PredRNN~\cite{wang2017predrnn,DBLP:journals/corr/abs-2103-09504}, PredRNN++~\cite{wang2018predrnn++}, MIM~\cite{wang2019memory}, E3D-LSTM~\cite{DBLP:conf/iclr/WangJYLLF19}, SA-ConvLSTM~\cite{lin2020self}. These methods have achieved remarkable results in ST-PL.

LSTM based models are mainstream in ST-PL. However, the input and context of previous models are solely correlated by CNN layers and channel-wise addition operation. Hence, with the increase of models' depth, correlations between the current input and upper context will decline as information flows through layers. To improve the correlation and capture important parts of input and context, we design \textbf{C}ontext \textbf{E}mbedding (CE) block to re-weight input and context states in an iteratively interacted mode to reflect the spatiotemporal details. CE block utilizes lightweight CNN layers to iteratively focus on important parts for subsequent prediction to enhance the correlations and capture the variation trend.

Spatiotemporal sequences contain complex semantic features, whereas the certainty of frames is exceptionally fuzzy. Previous work predicts increasingly blur details because they do not balance focusing on the changed regions and weakening the expression on unchanged parts. Instead, they merely concentrate on global spatiotemporal flows of given frames in hidden states, resulting in more extra params and ignorance of fine-grained variations. We creatively divide latent states into multi-scales to capture details of specific regions in parallel by proposing Multi-Scale \textbf{S}patiotemporal \textbf{E}xpression (SE) block. SE block captures fine-grained details based on the self-attention mechanism, which improves the dominant changed regions well expressed and simultaneously weakens the negligible parts with lower expression.

We integrate CE and SE blocks by proposing \textbf{C}ontext Embedding and \textbf{M}ulti-Scale \textbf{S}patiotemporal Expression LSTM (CMS-LSTM), an extension structure of ConvLSTM to improve prediction quality especially the details. CMS-LSTM overcomes deficiencies of the isolated relationship of context and input and pays more attention to multi-scale spatiotemporal flows. The main contributions are as follows:

\begin{itemize}[leftmargin=*,nosep,nolistsep]
\item We design CE block and SE block to capture fine-grained details to promote prediction quality. CE block maintains consistency and extracts further correlations between the current input and upper context. SE block facilitates multi-scale dominant spatiotemporal flows’ expression and weakens the negligible parts simultaneously.
\item To the best of our knowledge, the proposed CMS-LSTM is the first to innovatively integrate context interaction enhancement and multi-scale spatiotemporal expression mechanism for detailed prediction. It achieves significant improvement and state-of-the-art results in numbers of metrics on two representative benchmarks and scenarios.
\item Qualitative and quantitative experiments have demonstrated the importance of context interactions and multi-scale spatiotemporal flows in ST-PL. The proposed CE block and SE block have the portability to transplant in other models.
\end{itemize}
\section{Methods}

\subsection{Overview of CMS-LSTM}
Considering the limitation of ConvLSTM, the core goals of CMS-LSTM are to maintain the spatiotemporal consistency and correlations among frames in LSTM layers, facilitate multi-scale dominant spatiotemporal flows’ expression and weaken the negligible ones simultaneously. In specific, CMS-LSTM is constructed by taking both considerations of context interactions and multi-scale spatiotemporal flows. 
\begin{figure}[h!]
  \centering
  \includegraphics[width=1\linewidth]{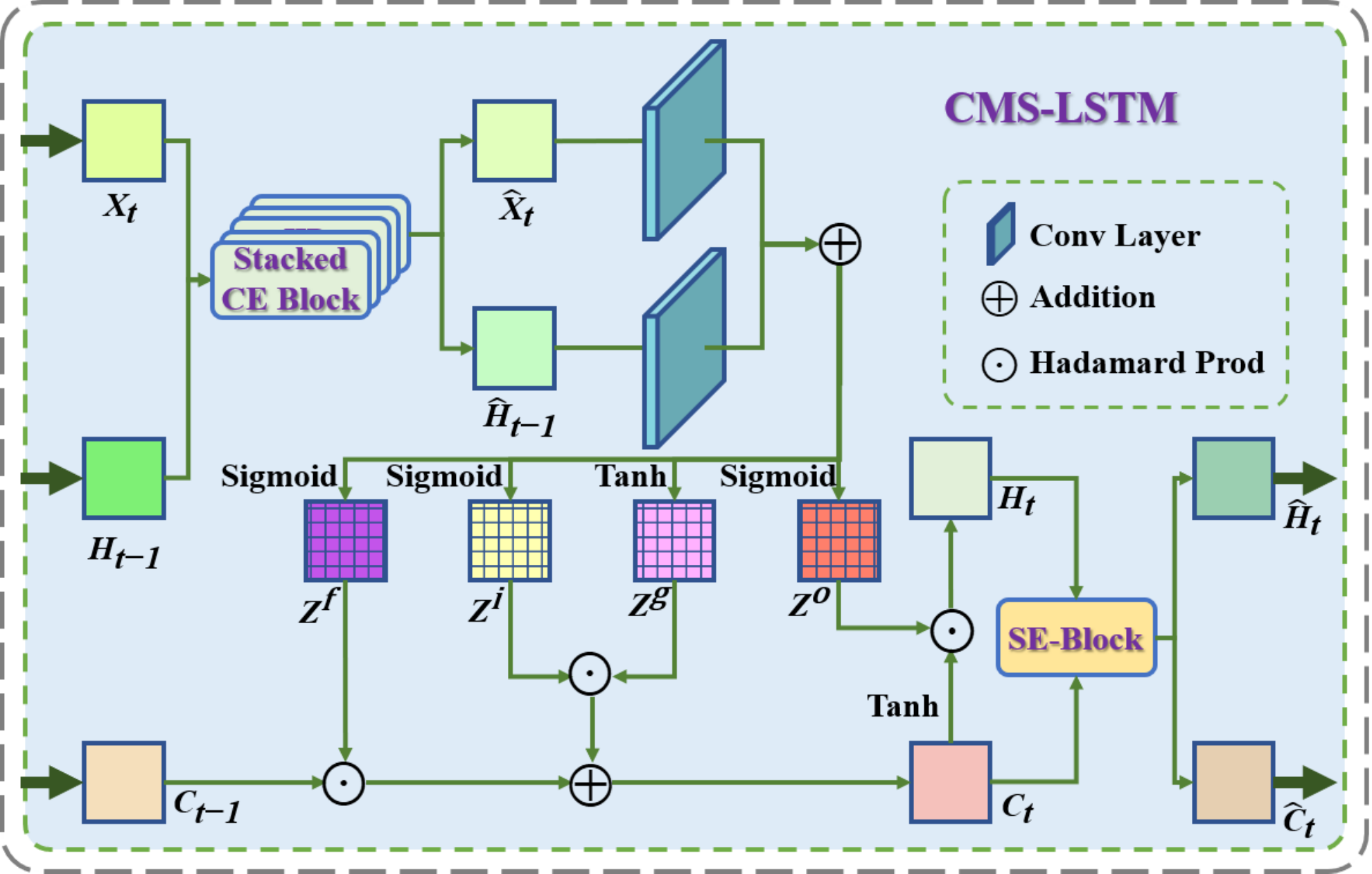}\vspace{-8pt}
  \caption{Architecture of CMS-LSTM. $H_{t-1}$ and $C_{t-1}$ represent output state and memory state of $t-1$ time respectively, $X_t$ represents the $t$ time input. $\hat{H}_t$ and $\hat{C}_t$ are the output of CMS-LSTM, i.e., the output state and memory state of $t$ time.}
  \label{fig::cesflstm}
  \vspace{-10pt}
\end{figure}

The architecture of proposed CMS-LSTM is illustrated in Fig.~\ref{fig::cesflstm}. Formally, the calculation process of CMS-LSTM can be expressed as follows:
\begin{equation}
    \begin{split}
    &\hat{X}_t,\hat{H}_{t-1}=CE(\cdots CE(X_t,H_{t-1})) \\
    & g_t=\tanh (W_{xg} \star \hat{X}_t+W_{hg} \star \hat{H}_{t-1}+b_g)\\
    &i_t=\sigma (W_{xi} \star \hat{X_t} + W_{hi} \star \hat{H}_{t-1}+b_i)\\
    &f_t=\sigma(W_{xf} \star \hat{X_t}+W_{hf} \star \hat{H}_{t-1}+b_f) \\
    &C_t=f_t\circ C_{t-1}+i_t\circ g_t\\
    & o_t=\sigma(W_{xo} \star \hat{X}_t +W_{ho}\star \hat{H}_{t-1}+b_o)\\
    &H_t=o_t\circ \tanh (C_t) \\
    &\hat{H}_t,\hat{C}_t=SE(H_t,C_t)
    \end{split}
    \label{equ:cesflstm}
\end{equation}
In Eq.~\ref{equ:cesflstm}. $\hat{X}_t$ and $\hat{H}_{t-1}$ represent the output of $n$ stacked CE blocks with intensive context interactions. Then, $H_t$ and $C_t$ are obtained through LSTM gate operations, which merely contain limited global spatiotemporal flows at present. We thus adopt $3$-scale SE block to extract multi-scale features for further spatiotemporal flows among neighbors, to obtain the final output $\hat{H}_t$ and $\hat{C}_t$ of CMS-LSTM. The structure of CE and SE blocks will be introduced later.

\subsection{CE Block} \label{subsec::CE-Block}

Rethinking the process of ConvLSTM~\cite{shi2015convolutional}, the input state $X_t$ and previous output state $H_{t-1}$ only interact separately by a CNN layer and addition operation. Limited interaction between the two states is crucial for the model's information loss and blurry prediction results. When the two states are completely independently entering into subsequent LSTM parts, correlations between the current input and upper context is bound to disappear as models become increasingly complex.

\begin{figure}[h!]
  \centering
  \includegraphics[width=\linewidth]{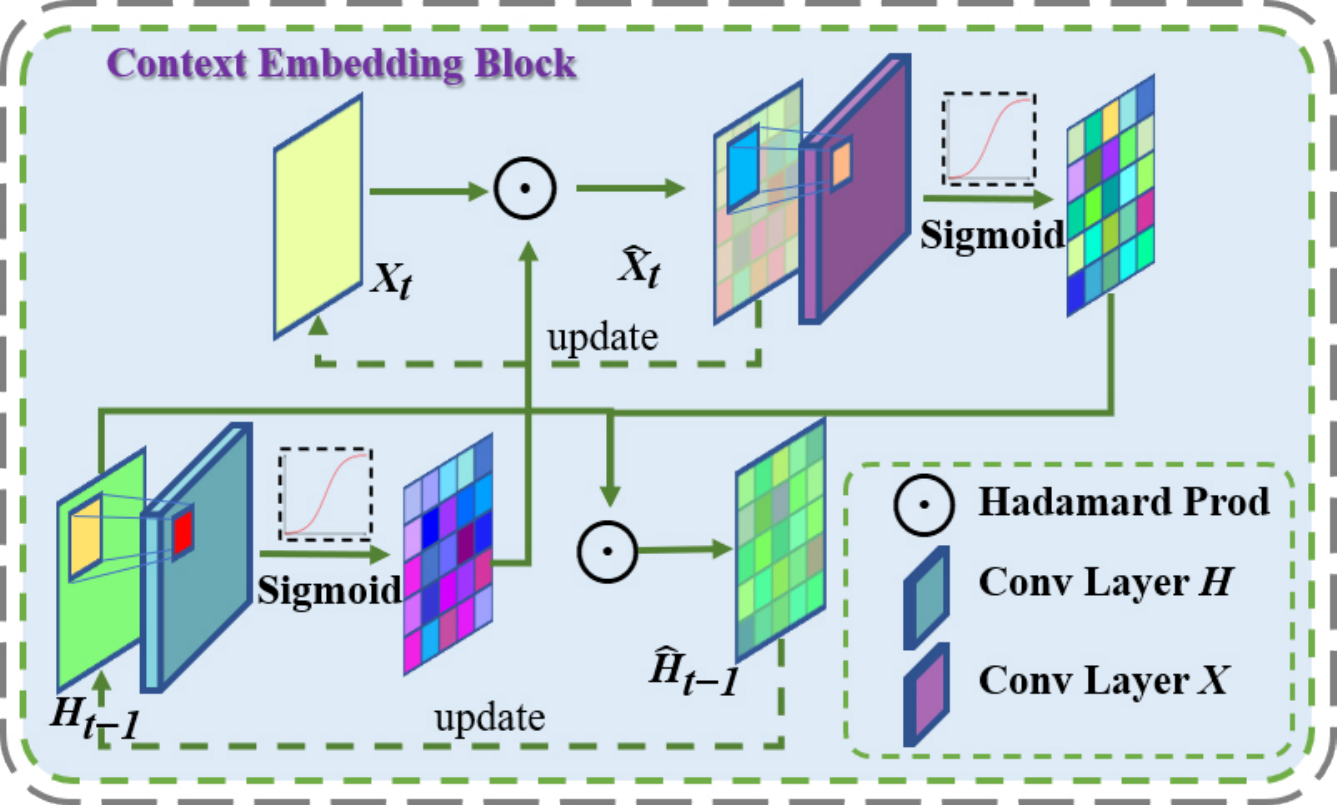}\vspace{-8pt}
  \caption{The pipeline of proposed CE block, where $H_{t-1}, X_t$ are the previous state and current input, respectively. $H$ and $X$ are $5 \times 5$ convolution layers to extract features of $H_t$ and $\hat{X}_t$, respectively. $\hat{H}_{t-1}$ and $\hat{X}_t$ are the output of CE block, representing the previous state and current input after context embedding, respectively.}
  \label{fig::hdblock}
  \vspace{-10pt}
\end{figure}

On the top of aforementioned, the current ConvLSTM and its extensions are incapable of re-weighting or capturing the important parts for the next timestamp. Therefore, we design CE block that contains additional operations to make persistent correlations of $X_t$ and $H_{t-1}$, to minimize the correlation decrease passing through LSTM layers and achieve a lasting relationship for better predicting performance. To achieve this, we utilize spatiotemporal features to generate a context weight map to enforce important information well captured, i.e., to concentrate on the changed parts while simultaneously weakening the fixed parts. Specifically, CE block (see Fig.~\ref{fig::hdblock}) consists of the following steps:
\begin{figure*}[h!]
  \centering
  \includegraphics[width=1\linewidth]{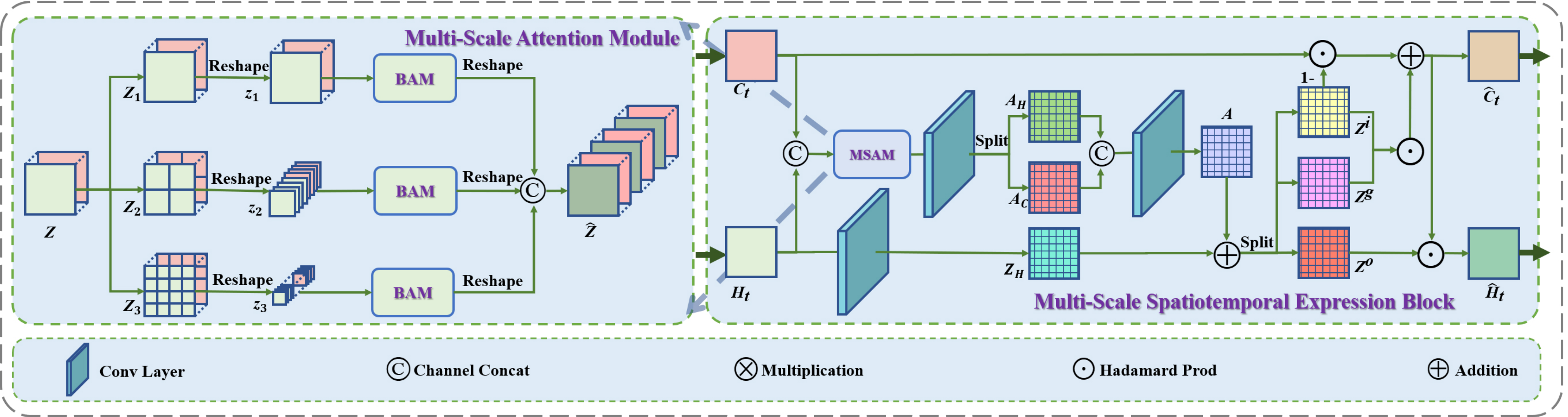}\vspace{-8pt}
  \caption{The pipeline of proposed SE block, with sub-module in the left part named \textbf{M}ulti-\textbf{S}cale \textbf{A}ttention \textbf{M}odule (MSAM), where $H_t$ and $C_t$ represent the output of original ConvLSTM, and $5 \times 5$ convolution layers are used to extract features, $\hat{H}_t$ and $\hat{C}_t$ represent the output state and memory state after multi-scale spatiotemporal flows' extraction.}
  \label{fig::lfblock}
  \vspace{-15pt}
\end{figure*}

\textbf{Step1.} 
To capture the important parts of current input that are helpful for long-term prediction, we generate a weight map of upper context via a $5\times 5$ kernel CNN layer to capture the context features, which indicates the potential movement trend in the following time stamp. Then, we adopt \textit{Sigmoid} function to normalize the weight map into $(0,1)$, and re-weight the input feature $X_t$ by the Hadamard product to highlight the important part of the input state. Finally, we multiply the weight map by a constant scale factor $s$ to avoid getting increasingly smaller as CE blocks stacked.

\textbf{Step2.} 
To enforce the context feature well absorb in the changing trend of current input frames, i.e., consider the current motion and weaken the unchanged parts with lower expression. We update $H_{t-1}$ by multiplying an input-related weight map to extract the subsequent motion concentration for subsequent prediction by the Hadamard product. The weight map has the same generation mode in Step 1, i.e., capture the local context motion features by a CNN layer and activation function with scale factor. Then, the updated context state $\hat{H}_{t-1}$ and input state $\hat{X}_t$ is obtained.

Formally, context correlations are extracted by the interaction mode as Eq.~\ref{equ:hdblock} in the proposed CE block, where $W_H,W_X$ are CNN and $\sigma$ represents $Sigmoid$ operation.
\begin{equation}
  \begin{split}
  &\hat{X}_t=s\times \sigma(W_H \star H_{t-1}+b_H) \circ X_t \\
  &\hat{H}_{t-1}=s \times \sigma(W_X \star \hat{X}_t+b_X) \circ H_{t-1} 
  \end{split}
  \label{equ:hdblock}
\end{equation}
To achieve richer interactions and minimize the extra params, we use stacked weight-shared CE blocks to extract abundant correlation further.

\subsection{SE Block} \label{subsec::SF-Block}

We find it's common that the prediction results of models become increasingly blur especially in the edges and details. The reason is that previous LSTM-based approaches mainly concentrate on modeling global spatiotemporal features and flows, regardless of multi-scale neighbor features among sequences. We emphasize the insufficiency of previous work in multi-scale spatiotemporal flow extractions and construct SE block for maximizing extract multi-scale implicit spatiotemporal flows to overcome previous weakness.

Considering those aforementioned, we construct SE block to enable the output state $H_t$ and memory state $C_t$ to contain abundant fine-grained spatiotemporal information. The pipeline of SE block is illustrated in Fig.~\ref{fig::lfblock}, which can be summarized as the following two steps.

\textbf{Step1. Multi-Scale Spatiotemporal Features Expression} 
To extract fine-grained spatiotemporal features in latent states, we adopt self-attention mechanism to obtain local features of specific scales. To avoid additional computation load and params on two hidden states, we stack the spatiotemporal states $H_t, C_t\in \mathbb{R}^{C \times H \times W}$ into $Z \in \mathbb{R}^{C \times H \times W \times 2}$ to improve the parallel efficiency.

To capture fine-grained feature in specific scale, we divide $Z$ into $n$ multi-scale groups $\{Z_1,\cdots, Z_n\}$ according to segmentation rules $R = \{R_1, \cdots, R_n\}$, then each $Z_i,i\in [1,n]$ is stacked in $C$ channel to compose $\{z_1, \cdots, z_n\}$. For each $z_i,i\in [1,n]$, we use standard self-attention module to reflect the importance of representing spatiotemporal characteristics of different regions, and denote as:
\begin{equation}
  \begin{split}
  &Z_i=\mathop{\text{\textit{DIVIDE}}}_{R_i \in R}([H_t,C_t]) \quad i\in [1,n]\\
  &\hat{z}_i=\text{\textit{BAM}}(\mathop{\text{\textit{STACK}}}_{C} (Z_i)) \quad i \in [1,n]
  \end{split}
  \label{equ:part1-1}
\end{equation}
Where BAM is a standard self-attention operation, which uses $1\times 1$ CNN layers to map $z_i$ into $K,Q,V$ and obtains updated $\hat{z}_i$ as follows:
\begin{equation}
  \begin{split}
  \text{\textit{BAM}}(z_i) \triangleq \text{\textit{Softmax}}(Q^T \times K)\times V + z_i
  \end{split}
  \label{equ:selfatt}
\end{equation}
After that, the multi-scale latent features are restored in $H$ and $W$ channels to recover the original shape, then we concat these multi-scale features in $C$ channel to composing $\hat{Z} \in \mathbb{R}^{nC\times H \times 2W}$. Ultimately, feature maps $A_H,A_C \in \mathbb{R}^{nC \times H \times W}$ are calculated by $5 \times 5$ CNN layer taking $\hat{Z} \in \mathbb{R}^{nC \times H \times W \times 2}$ as input and separated in the last channel.
\begin{equation}
  \begin{split}
  &\hat{Z}=\mathop{\text{\textit{CONCAT}}}_{C}(\mathop{\text{\textit{RESTORE}}}_{R_i \in R} {(\hat{z}_i)}) \quad i\in [1,n]\\
  &[A_H,A_C]=\text{\textit{SEPARATE}}(W_Z \star \hat{Z} + b_z)
  \end{split}
  \label{equ:part1-2}
\end{equation}
We successfully obtain the fine-grained features $A_H$ and $A_C$. They consist of abundant spatiotemporal correlations and details of the previous two states.

\textbf{Step2. Spatiotemporal Implicit States Update}
We utilize $A_H,A_C$ to update latent states with abundant details. Specifically, we stack the spatiotemporal related latent feature into channel dim and follow by a $5\times 5$ CNN layers, then split into 3 parts: $Z^i, Z^g,$ and $Z^o$, respectively.
\begin{equation}
    \begin{split}
    &i_t=\sigma(W_{Ai}\star [A_H,A_C]+W_{hi}\star Z_H+b_i) \\
    &g_t=\tanh(W_{Ag}\star [A_H,A_C]+W_{hg}\star Z_H+b_g)\\
    &o_t=\sigma(W_{Ao}\star [A_H,A_C]+W_{ho}\star Z_H+b_o)
    \end{split}
    \label{equ:updatec}
\end{equation}
Then, the memory state $\hat{C}_t$ and output state $\hat{H}_t$ integrate the detailed multi-scale features and further updated as follows:
\begin{equation}
    \begin{split}
    &\hat{C}_t=(1-i_{t})\circ {C_t}+i_{t}\circ g_{t} \\
    &\hat{H}_{t}=o_t\circ \hat{C}_t
    \end{split}
    \label{equ:updateh}
\end{equation}
With the construction of SE block, memory state and output state focus more on the detailed sequence changes in the long-term prediction and can effectively counter the gradually fuzzy prediction results.

\section{Experiments}
\subsection{Implementation Details}
We use the same $4$-layer LSTM architecture with $64$ hidden states for fair comparisons. Setting mini-batch to $8$ and initial learning rate to $0.001$, scheduled sampling~\cite{bengio2015scheduled} and layer normalization~\cite{ba2016layer} are simultaneously adopted during training. We use $L_1 +L_2$ loss with AdamW~\cite{loshchilov2018fixing} optimizer to train the model, We set scale factor of CE as $s=2$.
\subsection{Datasets}

\textbf{Moving MNIST}
Moving MNIST~\cite{srivastava2015unsupervised} is a common benchmark in ST-PL, depicting $2$ digits’ movement with constant velocity. Each data contains $64 \times 64 \times 1$ consecutive frames with $10$ for input and $10$ for prediction, $10,000$ randomly generate sequences for training and $10,000$ fixed sequences for testing.

\textbf{Typhoon}
Typhoon dataset is distributed by CEReS~\cite{DBLP:journals/remotesensing/YamamotoIHT20}. We normalize the radar observation data into $[0,1]$, resize the image to $64\times64\times1$. Each frame represents meteorological observation in the past $1$ hour. We use the given $8$-hour observation data to predict the next $4$ hours, with $1809$ sequences for training and $603$ sequences for testing.

\subsection{Comparisons with SOTA Methods}
We compare the proposed model with previous SOTA methods quantitatively and qualitatively to demonstrate our method's advantages and effectiveness.

\begin{figure}[t!]
  \centering
  \begin{minipage}[ht]{1\linewidth}
  \centering
  \includegraphics[width=1\linewidth]{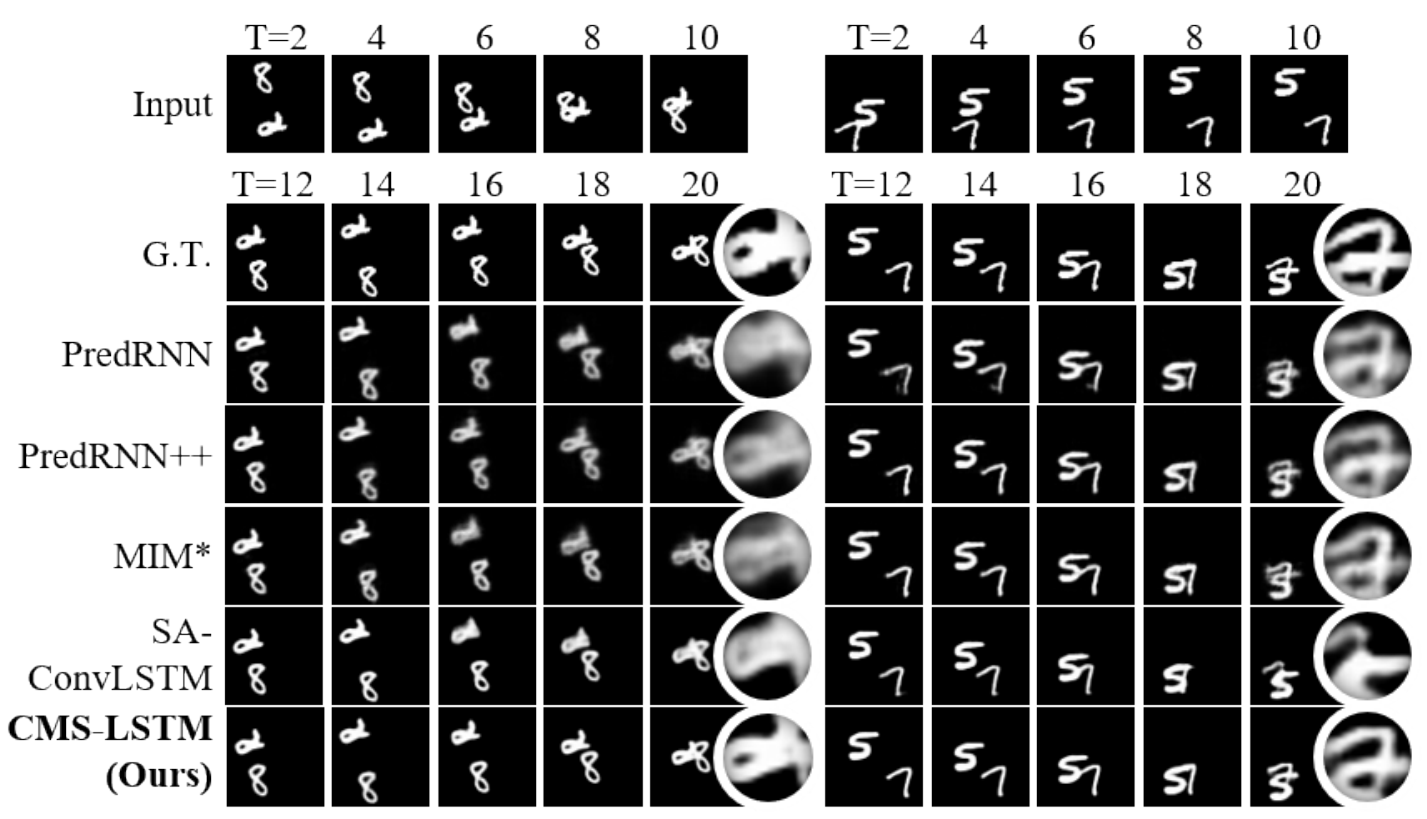}\vspace{-10pt}
  \caption{Qualitative comparisons of previous models on Moving MNIST test set at $80,000$ iterations. The output frames are shown at one-frame intervals. We magnify the local of prediction results for detailed comparisons at the last frame.}
  \label{fig::expmnist}
  \end{minipage}%
  \vspace{-10pt}
\end{figure}

\textbf{Results on Moving MNIST} \label{subsubsec::expmnist}
We set $80,000$ iterations consistent with~\cite{wang2017predrnn,lin2020self} and $400,000$ iterations for better performance. Quantitative and qualitative comparisons are shown in Tab.~\ref{tab::expmnist} and Fig.~\ref{fig::expmnist}, respectively. PSNR, SSIM, MSE, and MAE are used for quantitative comparisons. The performance improves as the SSIM and PSNR increase and the MSE and MAE decrease.
Results in Tab.~\ref{tab::expmnist} demonstrate the superiority of our method on Moving MNIST dataset in all above metrics, improving $14.5\%$ and $3.9\%$ on PSNR and SSIM, and reducing $41.7\%$ and $33.4\%$ on MSE and MAE respectively compared with SA-ConvLSTM~\cite{lin2020self}.

\begin{table}[t!]
  \caption{Quantitative comparisons of previous SOTA models on Moving MNIST test set. All models predict $10$ frames by observing $10$ previous frames. We also train $400,000$ iterations (CMS-LSTM*) for higher performance.}\vspace{-8pt}
  \resizebox{1\linewidth}{!}{%
  \centering
  \begin{tabular}{l|l|l|l|l|l}
  \toprule
  Models    & \#Params                 & PSNR $\uparrow$             & SSIM $\uparrow$              & MSE $\downarrow$             & MAE $\downarrow$             \\ \hline 
  DDPAE~\cite{DBLP:conf/nips/HsiehLHLN18} & - & 21.170  & 0.922 & 38.9  & 90.7 \\
  CrevNet~\cite{DBLP:conf/iclr/YuLEF20} & - &  \multicolumn{1}{l|}{-}& 0.928  & 38.5  & \multicolumn{1}{l}{-} \\
  PhyDNet~\cite{DBLP:conf/cvpr/GuenT20a} & - & 23.120  & 0.947  & 24.4 & 70.3  \\ 
  PDE-Driven~\cite{2021pdedriven} & - & 21.760 & 0.909  & \multicolumn{1}{l|}{-}&\multicolumn{1}{l}{-} \\ \hline
  PredRNN~\cite{wang2017predrnn}     & 13.799 M            & 19.603                & 0.867                  & 56.8                & 126.1              \\
  PredRNN++~\cite{wang2018predrnn++}    & 13.237 M            & 20.239              & 0.898        & 46.5               & 106.8             \\
  MIM*~\cite{wang2019memory}     & 27.971 M               & 20.678  & 0.910                  & 44.2             & 101.1                \\
  E3D-LSTM~\cite{DBLP:conf/iclr/WangJYLLF19}     & 38.696 M             & 20.590                & 0.910                  & 41.7                & 87.2                 \\
  SA-ConvLSTM~\cite{lin2020self}   & 10.471 M  & 20.500                & 0.913  & 43.9            & 94.7                 \\ \hline
  CMS-LSTM & 7.968 M & 21.955 & 0.931 & 33.6  & 73.1 \\
  \textbf{CMS-LSTM*} & \textbf{7.968 M} & \textbf{23.682} & \textbf{0.949} & \textbf{24.3} & \textbf{58.1} \\ 
  \bottomrule
  \end{tabular}
  }
  \label{tab::expmnist}
  \vspace{-15pt}
\end{table}

Results in Fig.~\ref{fig::expmnist} show that CMS-LSTM has better capability to capture variations over digits, especially deals with the trajectory of overlap digits and maintains the clarity over time. In contrast, predicted frames of other methods appear blurry in the digits and fail to deal with overlap digits.

\begin{figure*}[!htp]
  \centering
  \includegraphics[width=1\linewidth]{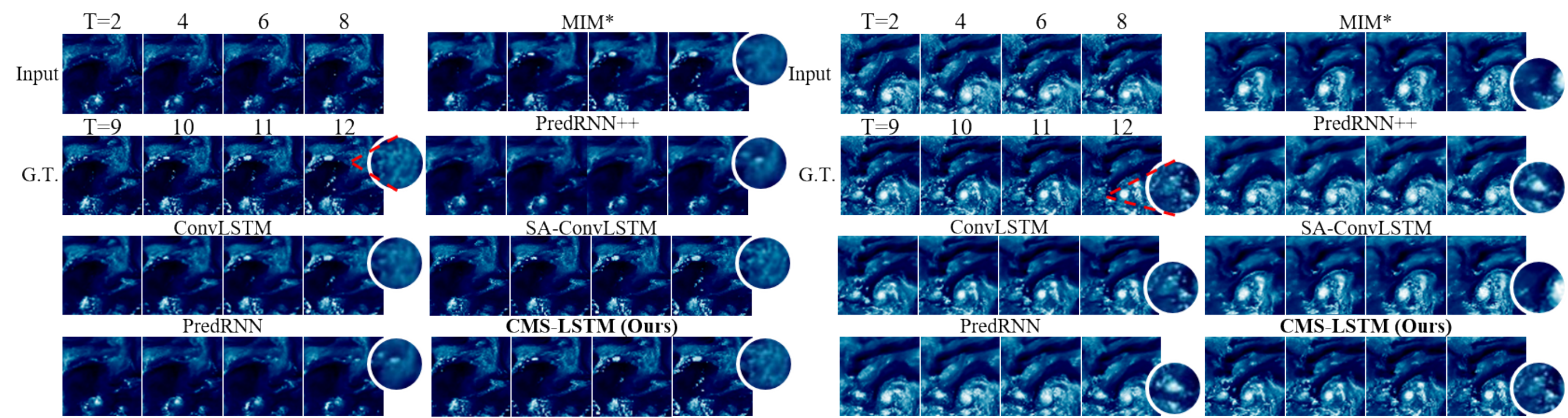}\vspace{-10pt}
  \caption{Qualitative comparisons of previous SOTA models on Typhoon test set. The output frames are shown at one-frame intervals. We frame the local of predicted results for additional detailed comparisons at the last frame.}
  \label{fig::exptyphoon}
		\vspace{-15pt}
\end{figure*}

\textbf{Results on Typhoon}
We train the proposed models for $100,000$ iterations and make fair comparisons with previous methods. PSNR, SSIM, MSE, and MAE are adopted to evaluate these models' performance qualitatively and quantitatively, corresponding to Fig.~\ref{fig::exptyphoon} and Tab.~\ref{tab::exptyphoon}.

\begin{table}[!ht]
\caption{Quantitative comparisons of previous SOTA models on Typhoon test set. All models predict the next $4$ frames (atmosphere trends for the next $4$ hours) via $8$ observed meteorological data. We also use $7\times 7$ convolution layers (denoted as CMS-LSTM\#) for higher performance.}\vspace{-8pt}
		\label{tab::exptyphoon}
\setlength{\tabcolsep}{10pt}
		\centering
\resizebox{\linewidth}{!}{%
  \begin{tabular}{l|l|l|l|l|l|l|l|l|l|l|l|l|l|l|l|l}
  \toprule
Models                                                & \multicolumn{4}{c|}{PSNR $\uparrow$}                  & \multicolumn{4}{c|}{SSIM $\uparrow$}                 & \multicolumn{4}{c|}{MSE $\downarrow$}               & \multicolumn{4}{c}{MAE $\downarrow$}               \\ \hline
ConvLSTM~\cite{shi2015convolutional} & \multicolumn{4}{l|}{26.353}                           & \multicolumn{4}{l|}{0.851}                           & \multicolumn{4}{l|}{10.43}                          & \multicolumn{4}{l}{119.6}                          \\
PredRNN~\cite{wang2017predrnn}      & \multicolumn{4}{l|}{27.637}                           & \multicolumn{4}{l|}{0.887}                           & \multicolumn{4}{l|}{7.71}                           & \multicolumn{4}{l}{107.3}                          \\
PredRNN++~\cite{wang2018predrnn++}   & \multicolumn{4}{l|}{28.287}                           & \multicolumn{4}{l|}{0.891}                           & \multicolumn{4}{l|}{6.72}                           & \multicolumn{4}{l}{114.5}                          \\
MIM*~\cite{wang2019memory}         & \multicolumn{4}{l|}{26.721}                           & \multicolumn{4}{l|}{0.893}                           & \multicolumn{4}{l|}{9.14}                           & \multicolumn{4}{l}{132.2}                          \\
SA-ConvLSTM~\cite{lin2020self}     & \multicolumn{4}{l|}{28.456}                           & \multicolumn{4}{l|}{0.898}                           & \multicolumn{4}{l|}{7.07}                           & \multicolumn{4}{l}{94.2}                           \\ \hline
CMS-LSTM                                              & \multicolumn{4}{l|}{28.891}                           & \multicolumn{4}{l|}{0.907}                           & \multicolumn{4}{l|}{6.24}                           & \multicolumn{4}{l}{86.4}                           \\
\textbf{CMS-LSTM\#}                  & \multicolumn{4}{l|}{\textbf{30.207}} & \multicolumn{4}{l|}{\textbf{0.921}} & \multicolumn{4}{l|}{\textbf{5.21}} & \multicolumn{4}{l}{\textbf{76.0}} \\
  \bottomrule
\end{tabular}}
		\vspace{-20pt}
\end{table}

The proposed method outperforms existing techniques quantitatively in Tab.~\ref{tab::exptyphoon} and qualitatively in Fig.~\ref{fig::exptyphoon}. CMS-LSTM is the only model that performs well in the detail texture of frames, i.e., it can preserve and predict the potential trend of meteorological information.

Results in Tab.~\ref{tab::exptyphoon} demonstrate the superiority of the proposed method, with better spatiotemporal expression and prediction results, which further proves the tremendous necessity of interactions among latent states in ST-PL.

\section{Ablation Studies}

\subsection{Weight Map Visualization}
\begin{figure}[t!]
\vspace{-5pt}
  \centering
  \includegraphics[width=1\linewidth]{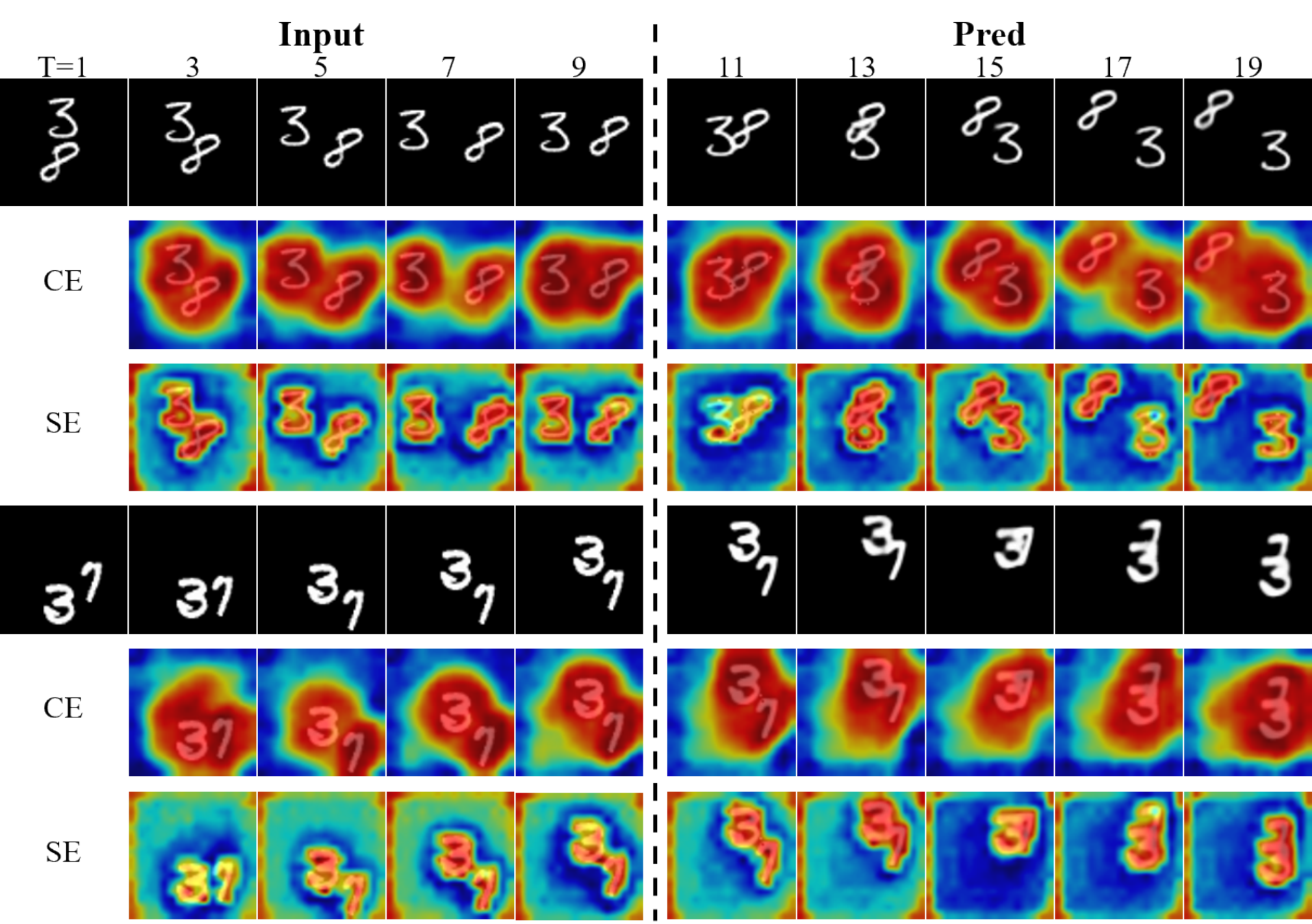}\vspace{-10pt}
  \caption{Visualization of the proposed blocks on Moving MNIST test set of the last LSTM layer, where the warm colors indicate higher weights. The proposed methods enable capturing important regions for better prediction.}
  \label{fig::attn_map}
		\vspace{-20pt}
\end{figure}

To illustrate the effectiveness of the proposed methods, we visualized the weight map calculated by CE and SE block in the last LSTM layer and randomly choose some examples from the test set of Moving MNIST as illustrated in Fig.~\ref{fig::attn_map}.

The weight map shows the important part of among frames. CE block enables important parts closely related to the context (e.g., moving trends) and input frames to be well captured, revealing the rough candidate regions as shown warmer color and keeping the unchanged parts with a lower weight. The output frames are closely related to the weight map where the important part is captured by SE block. SE block captures fine-grained details for prediction, which can alleviate the fuzzy texture in long-term prediction, especially in the challenge overlap cases. CMS-LSTM still accurately capture important parts and make satisfactory results.

As shown in Fig.~\ref{fig::attn_map}, CE block is effective in capturing a potential changed part, the larger weight that urges models to focus on these regions and ignore the fixed unchanged part, to avoid the wrong prediction results. The context embedding mode effectively predicts the potential trend of current input and upper context and effectively weakens the expression of irrelevant parts in the process of sequences. In SE block, the updated latent states are more focused on important parts, which helps the prediction results of the model to achieve higher prediction quality.

\begin{table*}[!htp]
    \caption{Ablation studies on Moving MNIST and Typhoon datasets. Models with and without CE block or SE block are tested sequentially in different backbones, as well as SE block with different scales in ConvLSTM. }\vspace{-8pt}
  \resizebox{\linewidth}{!}{%
  \begin{tabular}{l|ll|ll|ll|ll|ll|ll|ll|ll}
  \toprule
    \multicolumn{1}{c|}{\multirow{2}{*}{\textbf{Models}}} & \multicolumn{8}{c|}{\textbf{Moving MNIST}}                      & \multicolumn{8}{c}{\textbf{Typhoon}}         \\ \cline{2-17} 
    \multicolumn{1}{c|}{} &
      \textbf{PSNR $\uparrow$} &
      \textbf{$\Delta$} &
      \textbf{SSIM $\uparrow$} &
      \textbf{$\Delta$} &
      \textbf{MSE $\downarrow$} &
      \textbf{$\Delta$} &
      \textbf{MAE $\downarrow$} &
      \textbf{$\Delta$} &
      \textbf{PSNR $\uparrow$} &
      \textbf{$\Delta$} &
      \textbf{SSIM $\uparrow$} &
      \textbf{$\Delta$} &
      \textbf{MSE $\downarrow$} &
      \textbf{$\Delta$} &
      \textbf{MAE $\downarrow$} &
      \textbf{$\Delta$} \\ \hline
    ConvLSTM                                              & 18.523 & -      & 0.877 & -      & 70.4 & -     & 115.9 & -     & 26.353 & - & 0.851 & - & 10.4 & - & 119.6 & - \\
    w CE, w/o SE                                          & 21.189 & +2.666 & 0.918 & +0.041 & 39.1 & -31.3 & 82.8  & -33.1 & \textbf{29.022} & \textbf{+2.669}  & 0.906 & +0.055  & 6.32 & -4.08  & 87.0  & -32.6  \\
    w CE, w 1-scale SE                                    & 21.708 & +3.185 & 0.927 & +0.050 & 35.1 & -35.3 & 76.3  & -39.6 & 28.785 & +2.432  & 0.903 &  +0.052 & 6.58 & -3.82  & 89.3  &  -30.3 \\
    w CE, w 2-scale SE                                    & 21.858 & +3.335 & 0.929 & +0.052 & 33.8 & -36.6 & 74.3  & -41.6 & 28.650 & +2.297  & 0.901 & +0.050  & 6.75 &  -3.65 & 90.5  & -29.1  \\
    w/o CE, w SE                                          & 21.712 & +3.189 & 0.927 & +0.050 & 34.8 & -35.6 & 76.2  & -39.7 & 28.555 & +2.202  & 0.899 & +0.048  & 6.86 & -3.54  & 92.1  & -27.5  \\
    w CE, w SE                                            & \textbf{21.955} & \textbf{+3.432} & \textbf{0.931} & \textbf{+0.054} & \textbf{33.6} & \textbf{-36.8} & \textbf{73.1}  & \textbf{-42.8} & 28.891 & +2.538  & \textbf{0.907} &  \textbf{+0.056} & \textbf{6.24} & \textbf{-4.16}  & \textbf{86.4}  & \textbf{-33.2}  \\ \hline
    PredRNN     & 19.603 & -      & 0.867 & -      & 56.8 & -     & 126.1 & -     & 27.637 & -  & 0.887 &  - & 7.71 & -  & 107.3 &  - \\
    w CE, w/o SE            & 22.356 & +2.753 & 0.924 & +0.057 & 30.7 & -26.1 & 82.7  & -43.4 & 28.061       & +0.424  &  0.896     & +0.009  &   7.06   &  -0.65 &    102.7   & -4.60  \\
    w/o CE, w SE      & 22.761 & +3.158 & 0.931 & +0.064 & 28.7 & -28.1 & 76.9  & -49.2 &  28.516      & +0.879  &   0.900    & +0.013  &  6.51    & -1.20  &   \textbf{90.3}    & \textbf{-17.0}  \\
    w CE, w SE    & \textbf{23.210} & \textbf{+3.607} & \textbf{0.935} & \textbf{+0.068} & \textbf{26.3} & \textbf{-30.5} & \textbf{74.2}  & \textbf{-51.9} &  \textbf{28.864}      & \textbf{+1.227}  & \textbf{0.907}      &  \textbf{+0.020} &  \textbf{6.03}    &  \textbf{-1.68} &  94.7     & -12.6  \\ \hline
    SA-ConvLSTM                                         & 20.500 & -      & 0.913 & -      & 43.9 & -     & 94.7  & -     & 28.456 & -  & 0.898 &  - & 7.07 &  - & 94.2  & -  \\
    w CE, w/o SE                                          & \textbf{22.591} & \textbf{+2.091} & \textbf{0.929} & \textbf{+0.016} & \textbf{27.3} & \textbf{-16.6} & 79.0  & -15.7 & 28.628 &  +0.172 & 0.900 & +0.002  & 6.88 &  -0.19 & 89.3  & -4.90  \\
    w/o CE, w SE                                          & 21.700 & +1.200 & 0.928 & +0.015 & 34.8 & -9.10 & \textbf{75.7}  & \textbf{-19.0} &   28.690     & +0.234  &  0.903     &  +0.005 &   6.62   &  -0.45 &   89.4    &  -4.80 \\
    w CE, w SE                                           & 21.659 & +1.159 & 0.927 & +0.014 & 34.7 & -9.20 & 76.4  & -18.3 &  \textbf{29.505}     & \textbf{+1.049} &  \textbf{0.913}     & \textbf{+0.015}  &   \textbf{5.82}   & \textbf{-1.25}  &  \textbf{82.5}    &  \textbf{-11.7} \\ 
    \bottomrule  
  \end{tabular}%
  }
  \label{tab::ablmnist}
		\vspace{-15pt}
\end{table*}

\subsection{Ablation Study of CMS-LSTM}

We conduct ablation studies to verify the effectiveness of CE and SE block. Experiments below set $80,000$ ($100,000$) iterations in Moving MNIST (Typhoon) for training.

We verify the necessity of context interactions and multi-scale spatiotemporal flows by comparing CMS-LSTM removing CE and SE, respectively, and then using different scales to illustrate the necessity of the multi-scale spatiotemporal expression. The entire CMS-LSTM achieves the best performance compared with the original ConvLSTM. Comparing models with and without CE block demonstrates the necessity of context interactions. Moreover, experiments in multi-scale further show the importance of spatiotemporal flow extractions in different scales. 

Besides, to testify the portability of CE and SE, we transplant them into previous SOTA methods. Specifically, we compare PredRNN~\cite{wang2017predrnn} and SA-ConvLSTM~\cite{lin2020self} with/without CE block and SE block in the same experiment settings for quantitative comparisons on Moving MNIST and Typhoon dataset, results shown in Tab.~\ref{tab::ablmnist}.
Tab.~\ref{tab::ablmnist} further verifies the portability of the proposed blocks. With the transplant of them, previous models' performances do get significantly improved, indicating the ability of our methods to be transplanted in other spatiotemporal predictive models.

\section{Conclusions}
This paper creatively proposes effective modules named CE block and SE block focused on context interactions and multi-scale spatiotemporal expression, and then constructs CMS-LSTM. Qualitative and quantitative experiments demonstrate the superiority of our method dealing with uncertainty and overlap in sequences, showing state-of-the-art performance on representative datasets.

Ablation studies further verify the effectiveness and flexibility of our method. CE block can maintain the spatiotemporal consistency among long sequences, and SE block facilitates multi-scale dominant spatiotemporal flows’ expression and simultaneously weakens the negligible ones. Moreover, they can transplant to other spatiotemporal predictive related models to improve the performance markedly.

\section{Acknowledgment}
This work is supported by NSFC project Grant No. U1833101, SZSTI  Grant No. JCYJ20190809172201639, WDZC20200820200655001. Tsinghua \& Tencent Joint Research Laboratory.

\small{
\bibliographystyle{IEEEbib}
\bibliography{main}
}

\end{document}